\documentclass[sn-mathphys,Numbered]{sn-jnl}% Math and Physical Sciences Reference Style
\usepackage{graphicx}
\usepackage{xspace}
\usepackage{multirow}
\usepackage{amsmath,amssymb,amsfonts}
\usepackage{mathrsfs}
\usepackage[title]{appendix}
\usepackage{xcolor}
\usepackage{textcomp}
\usepackage{manyfoot}
\usepackage{booktabs}
\usepackage{algorithm}
\usepackage{algorithmicx}
\usepackage{algpseudocode}
\usepackage{listings}
\usepackage{multirow}
\usepackage{booktabs}
\usepackage{epstopdf}
\usepackage{hyperref}
\usepackage{natbib}
\usepackage{xspace} 
\begin{document}

\title[Article Title]{Comparative Evaluation of Digital and Analog Chest Radiographs to Identify Tuberculosis using Deep Learning Model}

%%=============================================================%%
%% Prefix	-> \pfx{Dr}
%% GivenName	-> \fnm{Joergen W.}
%% Particle	-> \spfx{van der} -> surname prefix
%% FamilyName	-> \sur{Ploeg}
%% Suffix	-> \sfx{IV}
%% NatureName	-> \tanm{Poet Laureate} -> Title after name
%% Degrees	-> \dgr{MSc, PhD}
%% \author*[1,2]{\pfx{Dr} \fnm{Joergen W.} \spfx{van der} \sur{Ploeg} \sfx{IV} \tanm{Poet Laureate} 
%%                 \dgr{MSc, PhD}}\email{iauthor@gmail.com}
%%=============================================================%%
\author[1]{\fnm{Subhankar} \sur{Chattoraj}}\email{subhankar.chattoraj@qure.ai}
\author*[1]{\fnm{Bhargava} \sur{Reddy}}\email{bhargava.reddy@qure.ai}
\author[1]{\fnm{Manoj} \sur{Tadepalli}}\email{manoj.tadepalli@qure.ai}
\author[1]{\fnm{Preetham} \sur{Putha}}\email{preetham.putha@qure.ai}
\affil[1]{\orgname{Qure.ai}, \orgaddress{\city{Mumbai}, \state{Maharashtra}, \country{India}}}

%%================================%%
%% Sample for structured abstract %%
%%================================%%

\abstract{\textbf{Purpose:} Chest X-ray (CXR) is an essential tool and one of the most prescribed imaging to detect pulmonary abnormalities, with a yearly estimate of over 2 billion imaging performed worldwide. However, the accurate and timely diagnosis of TB remains an unmet goal. The prevalence of TB is highest in low-middle-income countries, and the requirement of a portable, automated, and reliable solution is required. In this study, we compared the performance of DL-based devices on digital and analog CXR. The evaluated DL-based device can be used in resource-constraint settings.

\textbf{Methods:} A total of 10,000 CXR DICOMs(.dcm) and printed photos of the films acquired with three different cellular phones - Samsung S8, iPhone 8, and iPhone XS along with their radiological report were retrospectively collected from various sites across India from April 2020 to March 2021.

\textbf{Results:}10,000 chest X-rays were utilized to evaluate the DL-based device in identifying radiological signs of TB. The AUC of qXR for detecting signs of tuberculosis on the original DICOMs dataset was 0.928 with a sensitivity of 0.841 at a specificity of 0.806. At an optimal threshold, the difference in the AUC of three cellular smartphones with the original DICOMs is 0.024 (2.55\%), 0.048 (5.10\%), and 0.038 (1.91\%). The minimum difference demonstrates the robustness of the DL-based device in identifying radiological signs of TB in both digital and analog CXR.

\textbf{Conclusion:} 10,000 chest X-rays were utilized to evaluate the DL-based device in identifying radiological signs of TB. The AUC of qXR for detecting signs of tuberculosis on the original DICOMs dataset was 0.928 with a sensitivity of 0.841 at a specificity of 0.806. At an optimal threshold, the difference in the AUC of three cellular smartphones with the original DICOMs is 0.024 (2.55\%), 0.048 (5.10\%), and 0.038 (1.91\%). The minimum difference demonstrates the robustness of the DL-based device in identifying radiological signs of TB in both digital and analog CXR.}


%%\pacs[JEL Classification]{D8, H51}

%%\pacs[MSC Classification]{35A01, 65L10, 65L12, 65L20, 65L70}

\maketitle
\section{Introduction}
Chest X-ray (CXR) is an essential tool and one of the most prescribed imaging to detect pulmonary abnormalities, with a yearly estimate of over 2 billion imaging performed worldwide \cite{Intro_1}\cite{Intro_4}. Globally, the burden of people infected with Mycobacterium tuberculosis (TB) is one out of four people, and the chances of developing active tuberculosis (AT) among these individuals is 5\% - 10\% \cite{who_1} \cite{who_2}. Almost all AT cases occur in the 30 high-burden countries with low resources and stretched public health services \cite{Bhargava_1}. Recently due to the COVID-19 pandemic, it is estimated that 1.4 million fewer TB patients received timely care in 2020 than in the past year \cite{Intro_5}. However, with a large number of cases and already stretched medical facilities with the limited number of well-trained radiologists and the heavy workload in extensive healthcare facilities, timely diagnosis and management of TB is often a challenge \cite{Intro_6}. World Health Organization (WHO) has recommended using triage tests to determine the requirement of microbiological tests to diagnose TB as one of the potential solutions for managing TB in resource-limited settings \cite{Intro_7}. The triage tools can facilitate faster diagnosis in areas with limited trained personnel to interpret CXR. The increased burden of people with TB post-COVID-19 pandemic has necessitated a reliable automated, cost-effective pulmonary TB screening solution using CXR to ease the overburdened public health services.

The recent development of Deep Learning (DL) based in the multidisciplinary domain, such as medical imaging \cite{Intro_8}, computer vision \cite{Intro_9}, and signal processing \cite{Intro_10}, has been very promising. In a recent study, 2075 subjects in prison were prospectively evaluated for signs of TB. The reported specificity of the DL triage device was 74\% \cite{Intro_11}. In another study, multiple commercially available CXR-based diagnostic devices were used to identify radiological signs of TB in an active TB screening zone. The study recruited 23,954 patients, and only two products met the WHO Targeted Product Profile (TPP) (at $\geq 90$ specificity and $\geq 85$ sensitivity). Recently, in a study based on South African data of 165,754 CXR in 22,284 subjects which included 17\% of active TB cases, the deep learning system evaluated performed better than nine radiologists with a higher sensitivity of 13\% and specificity was inferior by 5\% \cite{Intro_17}. In a recent reader study based on the HIV population, an AI-assisted diagnostic assessment tool was evaluated on 683 subjects. The reader annotated the CXR with algorithm-assisted read and unassisted read, and the accuracy of detecting TB with the algorithm assistance was enhanced by 12\%. The standalone sensitivity and specificity of the algorithm were 0.67 and 0.87, respectively \cite{Intro_18}. Most studies evaluated DL-based computer-aided diagnostic (CAD) devices on digital CXR. However, the prevalence of TB was recorded highest in Asia (59\%) followed by Africa (26\%) \cite{Intro_19}, and in some remote locations, the availability of digital CXR is also a challenge. Recently, in a study, a DL-based CAD was evaluated on two publicly available databases (MIMIC-CXR and CheXpert). The CXR were images taken using a cellular smartphone. The proposed DL model achieved an overall AUC of 0.80. However, the goal of a reliable computer-aided diagnostic (CAD) model is to work efficiently in the remotest and rural areas that can efficiently provide acceptable performance on analog CXR (CXR photos taken from Cellular smartphone) comparison to digital CXR is still a challenge. Incorporating CAD devices capable of interpreting TB from analog CXR will provide healthcare workers with radiologist-level expertise. In numerous studies, the competency of CAD-based devices to detect radiological signs of TB has been reported \cite{Intro_12}. However, studies based on a comparative evaluation to detect TB from analog and digital CXR of the same patients still need to be discovered. 

One of the commercially available deep learning-based AI algorithms qXR (Qure.ai Technologies, Mumbai, India) has been previously evaluated in determining radiological signs of tuberculosis on CXR \cite{Intro_13}, prospective study in identifying normal vs. abnormal CXR \cite{Intro_14}, identification of malignant nodule on CXR \cite{Intro_15}, and missed or mislabelled findings \cite{Intro_16}. In this retrospective comparative study, we aimed to evaluate a deep learning-based AI algorithm (qXR) using standard evaluation metrics to identify radiological signs of TB on digital CXR and analog CXR. However, to the best of our knowledge, no comparative clinical study was conducted to detect TB from digital and analog CXR of the same patients.
%%%%%%%%%%%%%%%%%%%%%%%%%%%%%%%%%%%%%%%%%%
\section{Materials and Methods}
A total of 10,000 CXR DICOMs(.dcm), along with their radiological report, were prospectively collected from different sites across India from April 2020 to March 2021. Out of which, 1368 were from patients with GeneXpert positive, 2863 were from patients with GeneXpert negative, and 5768 were from patients with a completely normal CXR and no clinical signs of Tuberculosis. The cases were included in the final analysis per inclusion and exclusion criteria. The inclusion criteria were Posterior Anterior (PA) and Anterior-Posterior (AP) views, de-identified DICOMs format, and patient above 6 years of age where, as the exclusion criteria included incomplete metadata, non-chest radiographs, lateral view of the chest, CXR with postoperative defects and excessive motion artifacts.

%%%%%%%%%%%%%%%%%%%%%%%%%%%%%%%%%%%%%%%%%%%%
\begin{figure}[!htp]
\centering
\includegraphics[width=\textwidth]{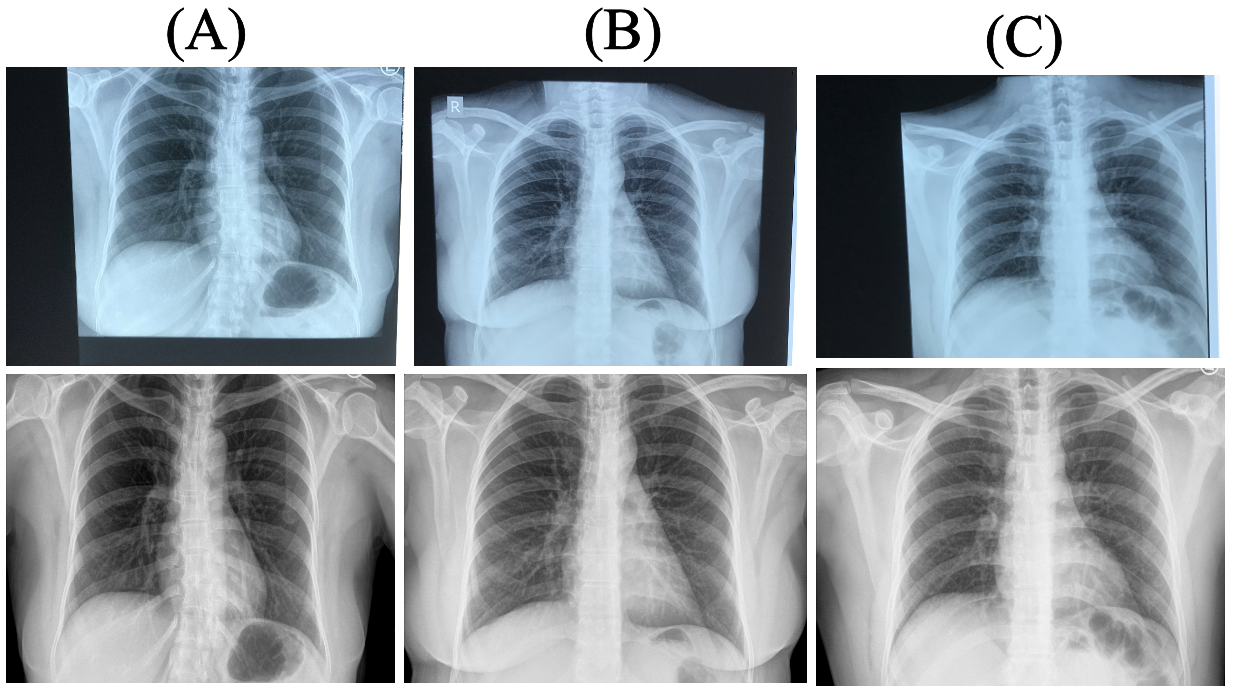}
\caption{The representative of the CXR images collected. In 1$^{st}$ row the (A), (B), (C) images captured using smartphone is given and in the 2$^{nd}$ row original CXR images are given.}
\label{data_sample}
\end{figure}
%%%%%%%%%%%%%%%%%%%

All these 10,000 CXR DICOMs were printed on films using a Fujifilm printer. A light box was set up with a tripod at a fixed place for mounting the mobile phone for standardized positioning. Photos of the films placed against the light-box were acquired with three different cellular phones - Samsung S8, iPhone 8, and iPhone XS. The CXR shows abnormalities by deep learning-based AI algorithm and is clinically diagnosed, confirming TB indications are advised CB-NAAT testing. The results from the tests were used as ground-truth to compare with the results provided by the DL-based AI algorithm using standard performance. In Fig. \ref{data_sample} the representative images of the data collected in given.
%%%%%%%%%%%%%%%%%%%%%%%%%%%%%%%%%%%%%%%%%%%%%%%
\begin{table}[!ht]
\centering
\caption{Detailed Distribution of the Evaluation Data}
\label{Overall_perform}
\begin{tabular}{@{}l|l@{}}
\midrule
\begin{tabular}[x]{@{}c@{}}\textbf{Findings}\end{tabular} & \begin{tabular}[x]{@{}c@{}}\textbf{Number of Images}\end{tabular} \\
\midrule
GeneXpert Positive&1,368\\
\midrule
GeneXpert Negative& 2,864 \\
\midrule
Normal &5,768\\
\midrule
\textbf{Total} &\textbf{10,000}\\
\end{tabular}
\end{table}
%%%%%%%%%%%%%%%%%%%%%%%%%%%%%%%%%%%%%%%%

%%%%%%%%%%%%%%%%%%%%%%%%%%%%%%%%%%%%%%%%%%%%%%%%%%%%%%%%%%
\subsection{Statistical analysis}
The performance of the DL-based device in categorizing CXR with radiological signs of TB on both digital and analog CXR was evaluated. The statistical analyses in terms of  AUC (Area under the curve), sensitivity, specificity, mean, and max score difference with the prediction on DICOMs, respectively, were performed using Python version 3.9.7 software.
%%%%%%%%%%%%%%%%%%%%%%%%%%%%%%%%%%%%%%%%%%%%%%%%%%%%%%%%%%
\section{Results}
The final study population resulted in 10,000 CXR from multiple centers across India. The median age of the final population was 39 years. The mean and interquartile patient age ranges were 41.36 and 22, respectively. There were 56.3\% male and 43.7\% female patients included in the study. The AUC of the DL-based device for detecting signs of tuberculosis on the original DICOM dataset was 0.941 (95\% CI: 0.907-0.975), with a sensitivity of 0.931 (95\% CI: 0.740-0.895) at a specificity of 0.826 (95\% CI: 0.782-0.904). As per standard practice for reporting metrics of continuously variable algorithms, the threshold at which the Youden index was maximum was considered for reporting the sensitivity and specificity across the datasets. The optimal thresholds were 0.51, 0.628, 0.593, and 0.639 for original DICOMs, iPhone 8 dataset, Samsung S8 dataset, and iPhone XS dataset, respectively. We noticed that the model predictions on DICOMs where higher compared to photos of films, necessitating a higher optimal and operating threshold.
%%%%%%%%%%%%%%%%%%%%%%%%%%%%%%%%%%%%%%%%%%%%
\begin{figure}[!htp]
\centering
\includegraphics[width=\textwidth]{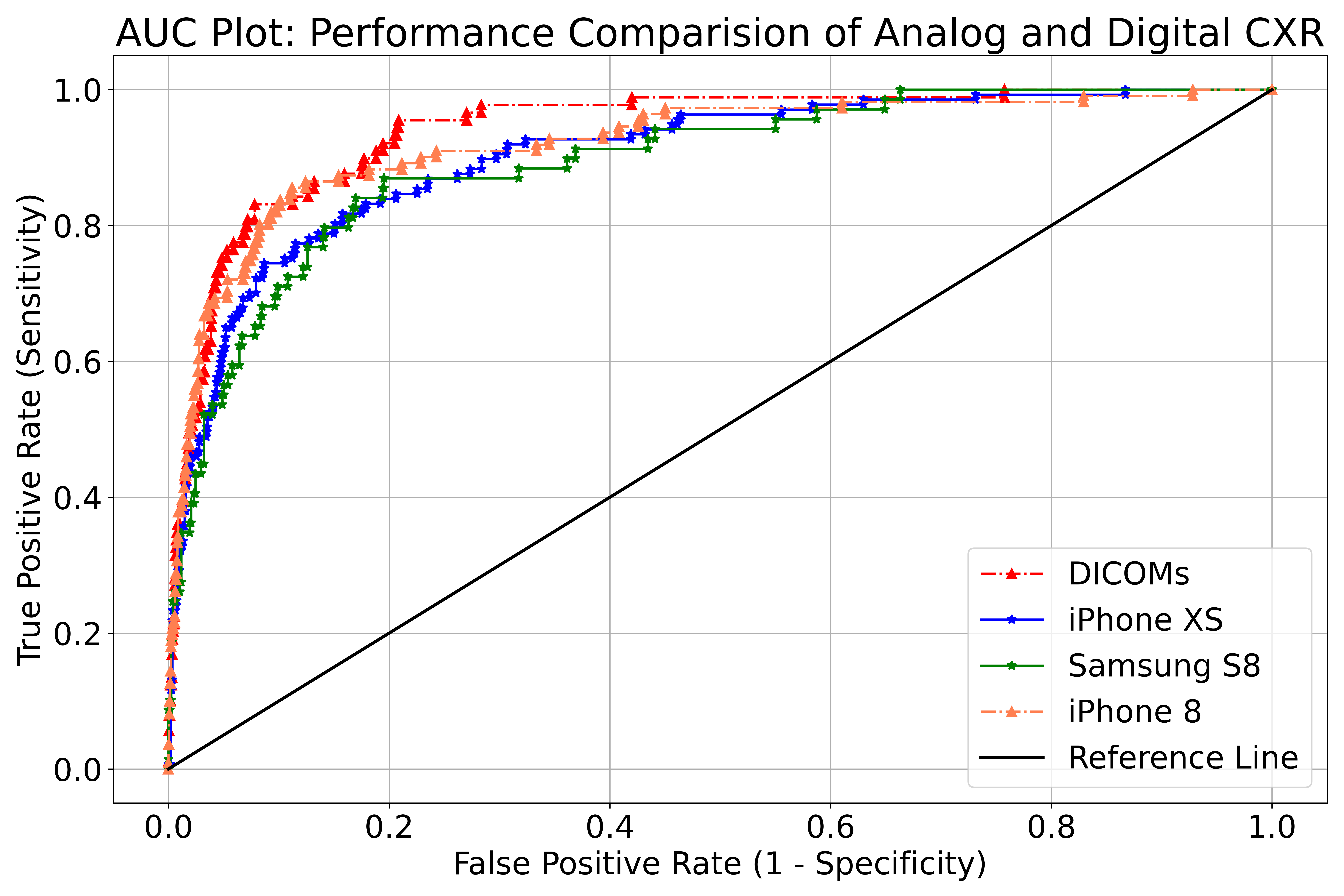}
\caption{Performance stratified for different cellular smart phones and DICOMs data}
\label{auc_plot_2}
\end{figure}
%%%%%%%%%%%%%%%%%%%

%%%%%%%%%%%%%%%%%%%%%%%%%%%%%%%%%%%%%%%%%%%%%%%
\begin{table}[!ht]
\centering
\caption{Overall Performance Evaluation of DL-based algorithm at Optimum Threshold; AUC: area under the curve; CSP = Cellular Smart Phone}
\label{results_binary}
% \scalebox{0.84}{
\begin{tabular}{@{}l|c|c|c|c|c@{}}
\midrule
\begin{tabular}[x]{@{}c@{}}\textbf{CSP}\end{tabular} &\multicolumn{3}{ c| }{\textbf{Performance Metrics (95\% CI)}}&\textbf{MeanSPD}& \textbf{MaxSPD}\\
\cmidrule{2-4}
&\textbf{AUC}&\textbf{Sensitivity }&
\textbf{Specificity}&& \\
\midrule
iphone 8 & \begin{tabular}[x]{@{}c@{}}0.917 \\(0.908 - 0.962)\end{tabular}& \begin{tabular}[x]{@{}c@{}}0.819 \\(0.790 - 0.86.5)\end{tabular}&\begin{tabular}[x]{@{}c@{}}0.794 \\(0.757 - 0.832)\end{tabular} &0.079 &0.308 \\
\midrule
Samsung S8& \begin{tabular}[x]{@{}c@{}}0.893 \\(0.843 - 0.944)\end{tabular} &\begin{tabular}[x]{@{}c@{}}0.829 \\(0.770 - 0.929)\end{tabular}  &\begin{tabular}[x]{@{}c@{}}0.804 \\(0.775 - 0.830)\end{tabular} &0.092 &0.340\\
\midrule
iPhone XS &\begin{tabular}[x]{@{}c@{}}0.903\\(0.869 - 0.937)\end{tabular}&\begin{tabular}[x]{@{}c@{}}0.817 \\(0.744 - 0.873)\end{tabular}&\begin{tabular}[x]{@{}c@{}}0.814 \\(0.793 - 0.827)\end{tabular}&0.104&0.298\\
\midrule
Digital CXR& \begin{tabular}[x]{@{}c@{}}0.941 \\(0.907 - 0.975)\end{tabular} &\begin{tabular}[x]{@{}c@{}}0.831 \\(0.740 - 0.895)\end{tabular} &\begin{tabular}[x]{@{}c@{}}0.826 \\(0.782 - 0.904)\end{tabular} &N.A.&N.A.\\
\midrule
\end{tabular}
% \vspace{-5mm}
\end{table}
%%%%%%%%%%%%%%%%%%%%%%%%%%%%%%%%%%%%%%%%

%%%%%%%%%%%%%%%%%%%%%%%%%%%%%%%%%%%%%%%%%%%%%%%
\begin{table}[!ht]
\centering
\caption{Stratifed Performance Evaluation of DL-based algorithm at Optimum Threshold; AUC: area under the curve; CSP = Cellular Smart Phone}
\label{stratified_per}
% \scalebox{0.84}{
\begin{tabular}{@{}l|c|c|c|c|c@{}}
\midrule
\begin{tabular}[x]{@{}c@{}}\textbf{Attributes}\end{tabular} &Subgroups&CXR Type &\multicolumn{3}{ c }{\textbf{Performance Metrics (95\% CI)}}\\
\cmidrule{4-6}
&&&\textbf{AUC}&\textbf{Sensitivity }&\textbf{Specificity}\\
\midrule
Age & 18 - 45&Digital&\begin{tabular}[x]{@{}c@{}}0.917 \\(0.908 - 0.962)\end{tabular}& \begin{tabular}[x]{@{}c@{}}0.819 \\(0.790 - 0.86.5)\end{tabular}&\begin{tabular}[x]{@{}c@{}}0.794 \\(0.757 - 0.832)\end{tabular}\\
\cmidrule{3-6}
& &Analog&\begin{tabular}[x]{@{}c@{}}0.912 \\(0.897 – 0.958)\end{tabular}& \begin{tabular}[x]{@{}c@{}}0.821 \\(0.789 – 0.863)\end{tabular}&\begin{tabular}[x]{@{}c@{}}0.813 \\(0.772 – 0.847) \end{tabular}\\
\midrule
& $\geq$ 46&Digital &\begin{tabular}[x]{@{}c@{}}0.926 \\(0.913 - 0.957)\end{tabular}& \begin{tabular}[x]{@{}c@{}}0.837 \\(0.816 - 0.872)\end{tabular}&\begin{tabular}[x]{@{}c@{}}0.795 \\(0.772 - 0.858)\end{tabular}\\
\cmidrule{3-6}

& &Analog &\begin{tabular}[x]{@{}c@{}}0.903 \\(0.872 - 0.927)\end{tabular}& \begin{tabular}[x]{@{}c@{}}0.821 \\(0.801 - 0.859)\end{tabular}&\begin{tabular}[x]{@{}c@{}}0.789 \\(0.762 - 0.816)\end{tabular}\\
\midrule
Gender & Male &Digital&\begin{tabular}[x]{@{}c@{}}0.934 \\(0.919 - 0.957)\end{tabular}& \begin{tabular}[x]{@{}c@{}}0.827 \\(0.812 - 0.873)\end{tabular}&\begin{tabular}[x]{@{}c@{}}0.814 \\(0.798 - 0.865)\end{tabular}\\
\cmidrule{3-6}
& &Analog&\begin{tabular}[x]{@{}c@{}}0.903 \\(0.892 - 0.959)\end{tabular}& \begin{tabular}[x]{@{}c@{}}0.802 \\(0.787 - 0.859)\end{tabular}&\begin{tabular}[x]{@{}c@{}}0.781 \\(0.768 - 0.839)\end{tabular}\\
\midrule
& Female &Digital &\begin{tabular}[x]{@{}c@{}}0.927 \\(0.892 - 0.951)\end{tabular}& \begin{tabular}[x]{@{}c@{}}0.811 \\(0.784 - 0.852)\end{tabular}&\begin{tabular}[x]{@{}c@{}}0.791 \\(0.749 - 0.828)\end{tabular}\\
\cmidrule{3-6}
& &Analog &\begin{tabular}[x]{@{}c@{}}0.912 \\(0.898 - 0.956)\end{tabular}& \begin{tabular}[x]{@{}c@{}}0.797 \\(0.782 - 0.862)\end{tabular}&\begin{tabular}[x]{@{}c@{}}0.782 \\(0.763 - 0.839)\end{tabular}\\
\midrule
\end{tabular}
% \vspace{-5mm}
\end{table}
%%%%%%%%%%%%%%%%%%%%%%%%%%%%%%%%%%%%%%%%
The AUC for TB detection was slightly lower on the dataset of photos of films, but there was no statistically significant difference in AUC across the cellular smartphone. Additionally, the mean and max difference in DL-based device scores for TB was computed for each cellular smartphone image dataset against the original DICOM dataset. In Table. \ref{Overall_perform}, detailed performance in terms of the performance evaluation metrics at optimal thresholds for the cellular smartphone datasets is given. The performance stratified by different cellular smartphones may help identify any differences in performance between the different smartphone models used in the study. In Figure. \ref{auc_plot_2}, the performance stratified by different cellular smartphones is represented. In Table. \ref{stratified_per}, stratified performance based on age (18-45), \& $\geq$ 46, and gender specific is detailed.
%%%%%%%%%%%%%%%%%%%%%%%%%%%
%%%%%%%%%%%%%%%%%%%%%%%%%%%%%%%%%%%%%%%%%%%%%%%%%%
\section{Discussion}
In this retrospective validation study, we collected 10,000 digital CXR, i.e., DICOMs, and analog CXR, i.e., CXR printed to film using a Fujifilm printer. Then the photo is taken after mounting the film on the light-box using the cellular smartphone from multiple healthcare facilities. The data collected was analyzed using commercially available deep learning-based AI algorithms (qXR). The purpose was to demonstrate the performance of the DL-based device in identifying TB and show the robustness of the DL-based device in identifying TB on both digital and analog CXR. The digital CXR achieved an AUC of 0.928, and the difference in AUC of digital and analog CXR captured from the different cellular devices were 0.0236 (2.69\%), 0.0283 (3.04\%), and 0.0178 (1.91\%), for iPhone 8, Samsung S8, and iPhone XS respectively.
Furthermore, the difference across multiple cellular smartphones was minimal, ranging from 1.9\% - 3\%, demonstrating the robustness of DL-based devices in identifying TB with minimum change in performance variations. A prior study reported that re-calibration of threshold is required for optimal performance of CAD devices \cite{Recab_1}. In this study, we have observed after the optimal selection of the threshold using the Youden index, the performance of the DL-based device improves significantly. 

Due to target domain divergence, most AI models need better generalization. Previously, evaluated the DL-based device, i.e., qXR, has been trained on more than 2.3 million CXR to identify multiple abnormalities on CXR including TB \cite{Intro_21}. The DL-based device has proven effective in retrospective and prospective settings to identify multiple pulmonary and extrapulmonary abnormalities. However, this is the first study conducted to demonstrate the efficacy of a DL-based device in identifying TB on both digital and analog CXR. The added advantage of our study from previously published literature \cite{Intro_20} is that the comparison of both analog and digital CXR from the same patient has been evaluated. The DL-based device considered in this study has the potential to be deployed in the remotest settings for the diagnosis of TB.

There were multiple limitations of our study. The retrospective study to evaluate DL-based devices has been based on data from a single demographic; thus, the reproducibility of similar performance needs to be validated. Secondly, the camera quality of the phone and lighting conditions can significantly influence the outcome of the DL-based device performance. The performance of the DL-based device in poor conditions also needs to be validated.
%%%%%%%%%%%%%%%%%%%%%%%%%%%%%%%%%%%%%%%%%%
\section{Conclusions}
The retrospective study demonstrated that DL-based devices can perform comparably to analog and digital CXR in identifying TB. The minimal difference observed across all the evaluated cellular smartphones reflects the robustness of the DL-based device. The study also demonstrated that the performance metrics obtained for different evaluation metrics were acceptable in clinical standards. This is the first study to assess the comparable performance of DL-based devices for both analog and digital CXR.

DL-aided device has the potential to be deployed in remote locations for TB screening. However, the study had some limitations, such as being retrospective and based on data from a single demographic. Furthermore, the quality of the cellular smartphone camera and lighting conditions can significantly influence the performance of the DL-based device, and its performance in poor conditions needs to be validated. Nonetheless, the study's findings suggest that DL-based devices can be an effective tool for TB screening, especially in resource-limited settings.
%%%%%%%%%%%%%%%%%%%%%%%%%%%%%%%%%%%%%%%%%%%%%%%%%%

\bibliography{sn-main}% common bib file
%% if required, the content of .bbl file can be included here once bbl is generated
%%\input sn-article.bbl

\end{document}